\documentclass{sigchi}

\usepackage{hhline,tabu}

\usepackage{color}

\usepackage{graphicx}

\usepackage{caption}
\usepackage{subcaption}

\usepackage{multirow}

\usepackage{verbatim}

\usepackage{balance}  
\usepackage{graphics} 
\usepackage{times}    
\usepackage{url}      
\usepackage{bm}
\usepackage[labelsep=period]{caption}

\makeatletter
\def\url@leostyle{%
  \@ifundefined{selectfont}{\def\UrlFont{\sf}}{\def\UrlFont{\small\bf\ttfamily}}}
\makeatother
\def\pprw{8.5in}
\def\pprh{11in}

\usepackage[pdftex]{hyperref}
\hypersetup{
pdftitle={SIGCHI Conference Proceedings Format},
pdfauthor={LaTeX},
pdfkeywords={SIGCHI, proceedings, archival format},
bookmarksnumbered,
pdfstartview={FitH},
colorlinks,
citecolor=black,
filecolor=black,
linkcolor=black,
urlcolor=black,
breaklinks=true,
}

\numberofauthors{1}
\author{
\alignauthor
Ming Zeng\textsuperscript{1}\thanks{equal contribution}, Haoxiang Gao\textsuperscript{1}\footnotemark[1], Tong Yu\textsuperscript{1}, Ole J. Mengshoel\textsuperscript{1},\\ Helge Langseth\textsuperscript{2}, Ian Lane\textsuperscript{1}, Xiaobing Liu\textsuperscript{3}\\
       \affaddr{\textsuperscript{1}Carnegie Mellon University, \{ming.zeng, haoxiang.gao, tong.yu, ole.mengshoel, ian.lane\}@sv.cmu.edu}\\
       \affaddr{\textsuperscript{2}The Norwegian University of Science and Technology, helge.langseth@ntnu.no}\\
       \affaddr{\textsuperscript{3}Google Brain, xbing@google.com}\\
}

\begin{document}

\hyphenation{para-digms represen-tative non-represen-tative}

\toappear{Permission to make digital or hard copies of all or part of this work for personal or classroom use is granted without fee provided that copies are not made or distributed for profit or commercial advantage and that copies bear this notice and the full citation on the first page. Copyrights for components of this work owned by others than ACM must be honored. Abstracting with credit is permitted. To copy otherwise, or republish, to post on servers or to redistribute to lists, requires prior specific permission and/or a fee. Request permissions from Permissions@acm.org.\\
{\emph{ISWC '18}}, October 8--12, 2018, Singapore, Singapore\\
\textcopyright 2018 Association for Computing Machinery.
\\
ACM ISBN 978-1-4503-5967-2/18/10\$15.00 \\
https://doi.org/10.1145/3267242.3267286
}

%

\title{Understanding and Improving Recurrent Networks for Human Activity Recognition by Continuous Attention}



\maketitle
\begin{abstract}

Deep neural networks, including recurrent networks, have been successfully applied to human activity recognition. Unfortunately, the final representation learned by recurrent networks might encode some noise (irrelevant signal components, unimportant sensor modalities, etc.). Besides, it is difficult to interpret the recurrent networks to gain insight into the models' behavior. To address these issues, we propose two attention models for human activity recognition: temporal attention and sensor attention. These two mechanisms adaptively focus on important signals and sensor modalities. To further improve the understandability and mean F1 score, we add continuity constraints, considering that continuous sensor signals are more robust than discrete ones. We evaluate the approaches on three datasets and obtain state-of-the-art results. Furthermore, qualitative analysis shows that the attention learned by the models agree well with human intuition.

\end{abstract}



\section{Introduction}
Wearable-based human activity recognition (HAR) systems are an indispensable component in mobile ubiquitous computing and human-computer interaction~\cite{bulling2014tutorial}. By integrating data from various sensor modalities (accelerometer, gyroscope, GPS, etc.), HAR systems are used in a large number of context aware applications such as daily life logging~\cite{chennuru2012mobile}, and cross-device user recognition~\cite{wang2017xrec}. To recognize users' activities, various machine learning algorithms have been engineered for specific application contexts~\cite{bao2004activity, bulling2014tutorial}.


Deep Neural Networks (DNNs), especially Recurrent Neural Networks (RNNs), are very good at discovering intricate structure in sequential data, and have proven their potential and pushed the state-of-the-art in HAR~\cite{hammerla2016deep,ordonez2016deep, guan2017ensembles}. A DNN consists of multiple layers of neurons and is built for automatic feature extraction, which reduces the need for designing hand-crafted features. Recently, one of the RNN models, so-called Long Short Term Memory (LSTM), has been very successfully employed in HAR~\cite{guan2017ensembles}. The LSTM encodes an input time series signals sequentially into a fixed length vector and then feeds it into a classifier. This sequential architecture is appealing, as it make it possible to capture long-range dependencies in a sequence. \\

\begin{figure}[tbph]
\centering
\includegraphics[width=0.5\textwidth]{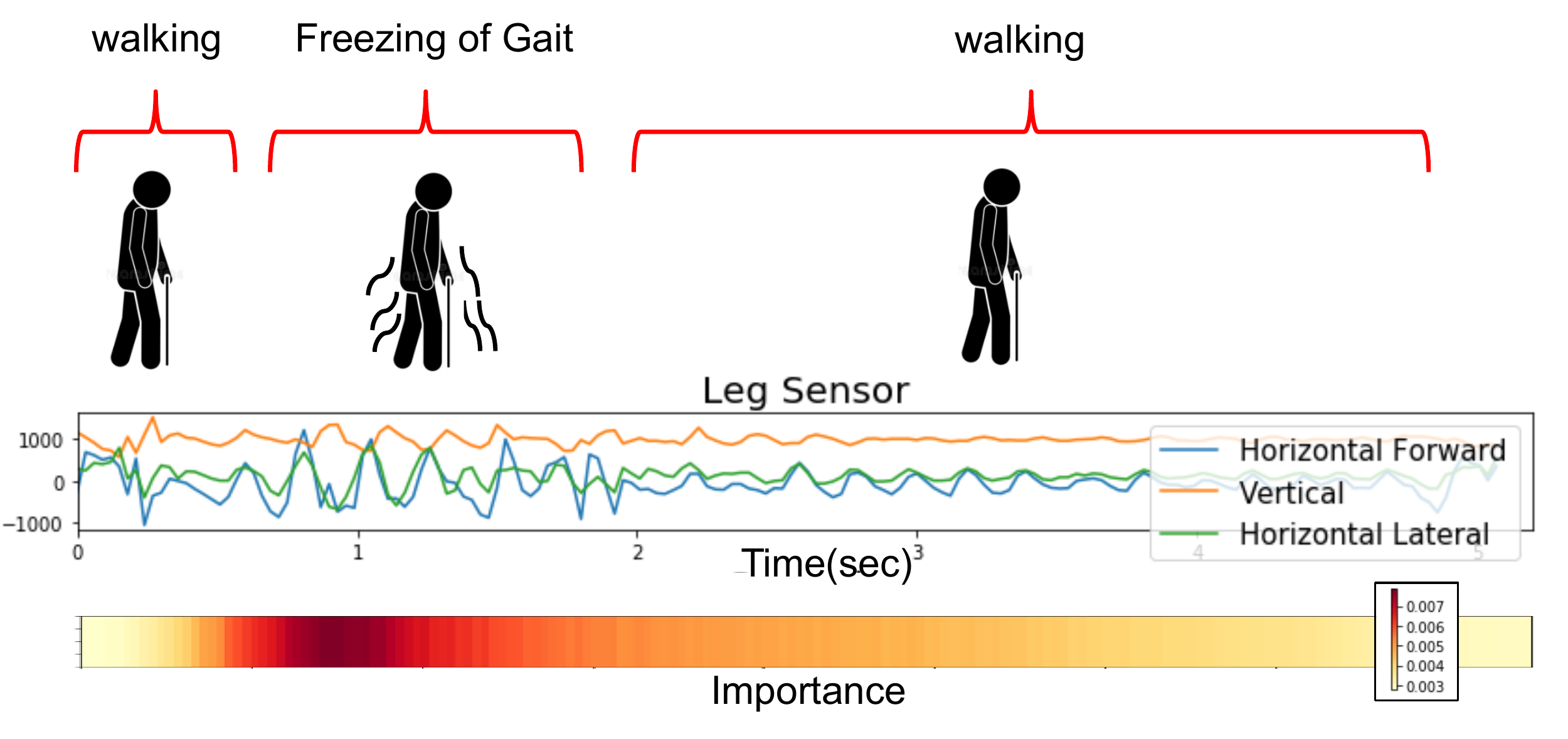}
\caption{\label{fig:fall_down} An example of a Freezing of Gait (FOG) detection for Parkinson disease from Daphnet Gait (DG) dataset~\protect\cite{bachlin2009potentials}. The important acceleration signal components for FOG is shown in dark red.}
\end{figure}


However, there are some challenges in HAR with LSTM: (i) The sensor signals may contain some irrelevant components. For example, the abnormal signal of the freezing of gait (FOG)~\cite{bachlin2009potentials} of Parkinson disease (Figure~\ref{fig:fall_down}) only shows up in a small time interval rather than in the entire window. Standard LSTM may not effectively detect FOG in this time series signal.
(ii) Different sensor modalities play different roles for recognizing different activities~\cite{zeng2014adaptive, yao2017deepsense}. Using unimportant sensors modalities may introduce substantial noise into HAR. For example, to detect FOG during walking, the accelerometer is more important than the magnetometer and gyroscope. And the sensor attached to the leg is more important than that on the trunk~\cite{bachlin2009potentials}.
(iii) The gains in good recognition results with RNNs have come at the cost of interpretability. It is desirable to understand the underlying basis for the decisions of neural models in HAR.    \\



To address the first challenge, noisy signals (i), we apply temporal attention LSTM to automatically ignore the unimportant parts in the input signals and highlight the important parts. To deal with the second challenge, sensor modality importance (ii), we propose to use sensor attention on the input layer to fuse the sensor modalities based on their importance. However, when applying attention through time, we notice that continuous signals are more robust than discrete ones. Thus, we propose a continuity constraint for the two attention models. Visualizing where in the input the models are attending enables us to understand the models' behavior.

The contributions of this paper are the following:
\begin{itemize}
\item
We propose temporal attention applied to the LSTM's hidden layer to highlight the important part of the time series signals.


\item
We propose sensor attention applied to the LSTM's input layer to reweight the sensor modality importance during training. 

\item
To further improve the two attention LSTMs on time series sensor data, we propose continuous attention constraints on both time attention and sensor attention by using additional regularizations.

\item 
By visualizing which parts of the input signals the models are attending to, we can get valuable insights into the models' behavior, thus improving interpreterbility.



\end{itemize}







\section{Related Work}
The standard way of dealing with HAR relies on hand-crafted features. Many existing feature extraction approaches use statistical features of raw signals, such as mean, variance, entropy, and correlation coefficients~\cite{bao2004activity}. 
Another branch of HAR feature extraction is transform coding, such as Fourier and wavelet transform~\cite{huynh2005analyzing}. 

Since designing hand-crafted features requires domain knowledge, 
it is desirable to develop a systematical feature learning approach to model the time series signals in HAR with multiple sensors~\cite{yang2015deep}. Deep neural networks (DNNs) has had revolutionary impact in speech recognition~\cite{chorowski2015attention}, image classification~\cite{xu2015show}, and machine translation~\cite{bahdanau2014neural}. DNNs also provided promising results in HAR domain ({\em e.g.,}~\cite{plotz2011feature,ronao2015deep, hammerla2015pd}). Deep Belief Network (DBN) is used for feature extraction, but fail to consider the order of signal and cannot outperform PCA-ECDF features~\cite{plotz2011feature}. Similar to speech recognition, the combination of traditional sequence models (HMMs) and DNNs is also applied in HAR~\cite{zhang2015human, alsheikh2016deep}. 
However, the fully-connected DNNs fail to consider the order of time series signals. The 1D convolutional neural netowrks (CNN) employ more effective signal processing units, such as convolution (capturing local dependency), pooling (capturing time invariant features), and it also makes use of the available label information in feature extraction~\cite{zeng2014convolutional}. To fuse different sensor modalities~\cite{yang2015deep, munzner2017cnn}, the 2D CNN regards the set of signals as an image. The width is the length of the signal and the height is the number of sensors. The image-like CNN not only captures the salient features, but also takes advantage of more sensors to obtain higher accuracy. Ord{\'o}{\~n}ez et al.~\cite{morales2016deep} demonstrate the use of transfer learning to reduce the influence on randomly initialized CNN weights. Recently, Zeng et al.~\cite{zeng2017semi} propose the CNN encoder-decoder and CNN ladder for HAR in the semi-supervised setting.

In order to capture the long-term information in time series data, the RNN with long short-term memory (LSTM) is proposed by Hochreiter and Schmidhuber~\cite{hochreiter1997long}, and successfully applied in HAR~\cite{ordonez2016deep, hammerla2016deep, guan2017ensembles}. Ord{\'o}{\~n}ez et al.~\cite{ordonez2016deep} combined convolutional and LSTM layers to provide promising results in recognition performance. Hammerla et al.~\cite{hammerla2016deep} compared various DNN models in HAR, including LSTMs, and CNNs. Guan et al.~\cite{guan2017ensembles} used ensemble LSTM model to capture diverse data during the training and significantly improved the recognition performance and robustness. However, a potential issue with LSTMs is that a neural network needs to be able to compress all the necessary information of a single input, which will involve noise or irrelevant parts. The attention approach is proposed to mitigate this problem for speech and natural language processing (NLP)~\cite{bahdanau2014neural, kim2017end}, and it offers means to explicate the inner workings of neural networks. Different from the speech and NLP, the dominant signals in HAR are continuous rather than discrete as discussed in the subsection Continuous Attentions. Sensors on different part of the body play different important roles for different activities. With this motivation, we built a new attention-based deep network architecture for HAR.


\section{Attention-based RNNs for Human Activity Recognition}

We frame HAR as a sequence classification problem~\cite{bao2004activity}. Given a sequence of sensor readings as input, namely $\mathbf{X}=(\mathbf{x}_1,...\mathbf{x}_T)$, where $T$ denotes the length of the signals, and $\mathbf{x}_t = [x_t^1,...,x_t^D]$ ($t \in \{1,...,T\}$) is a $D$ dimensional vector denoting a sensor reading at time $t$ for $D$ channels ({\em e.g.} acceleration has three channels: $x$-axis, $y$-axis, $z$-axis). The learning problem is to map the input sequence $\mathbf{X}$ to a target $y \in \mathbf{C}$, where $C=|\mathbf{C}|$ is the number of activity classes. We first introduce the baseline model, called long and short term memory (LSTM). The LSTM is also treated as a basic building block of our sequence classier in this paper.


\subsection{Standard LSTM}
\begin{figure}
\centering
\includegraphics[width=0.35\textwidth]{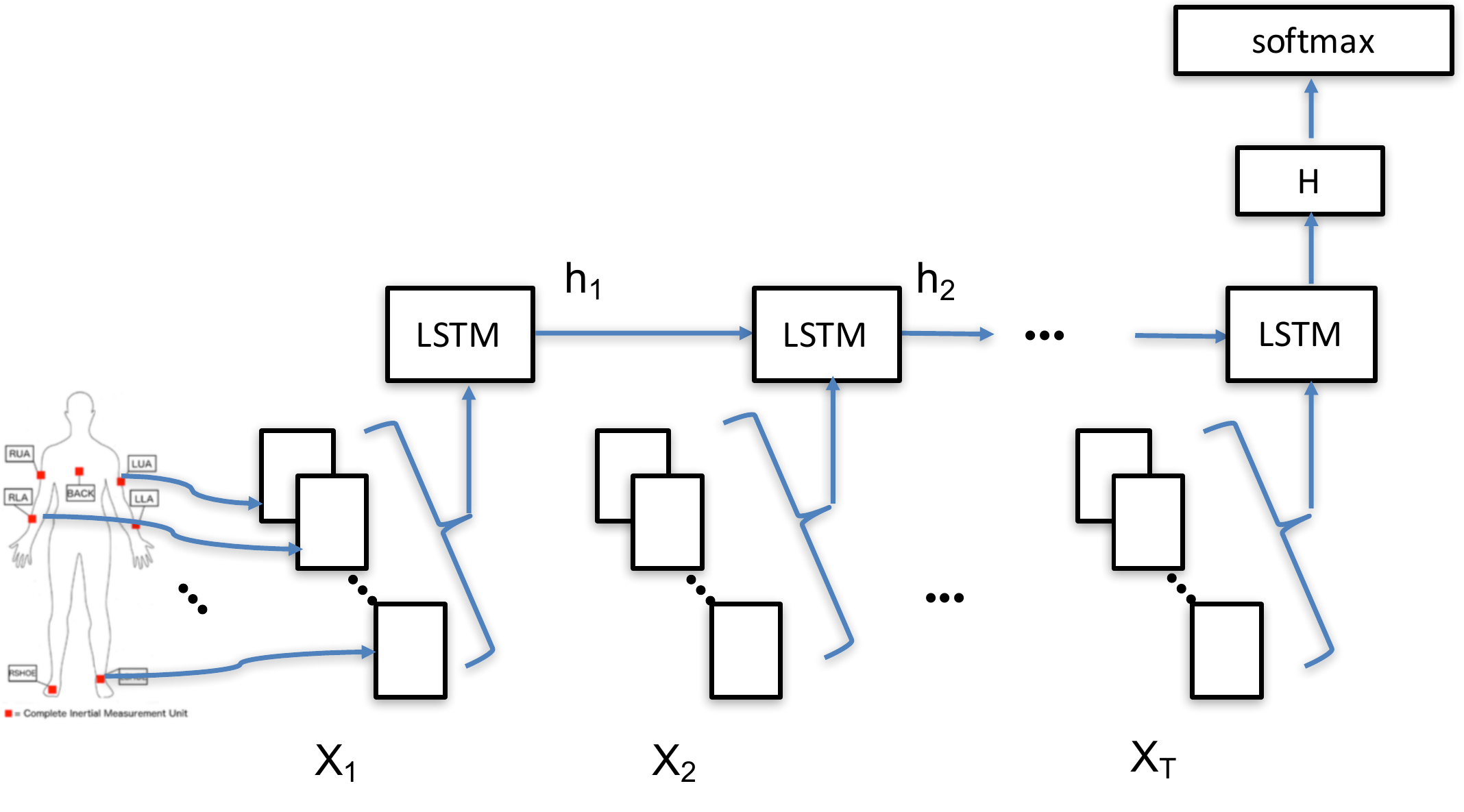}
\caption{\label{fig:lstm-structure} Structure of the standard LSTM~\protect\cite{hochreiter1997long}.}
\end{figure}

The LSTM is a recurrent network with four gates: $\mathbf{i}$ is the input gate, $\mathbf{f}$ is the forget gate, $\mathbf{o}$ is the output gate, and $\mathbf{c}$ is the cell activation vector~\cite{hochreiter1997long}. They can be described by the following equations:

\begin{align}
\mathbf{i}_t& = \sigma(\mathbf{x}_t \mathbf{W_{xi}} + \mathbf{h}_{t-1}\mathbf{W_{hi}} + \mathbf{b}_t) \label{lstm1} \\
\mathbf{f}_t& = \sigma(\mathbf{x}_t \mathbf{W_{xf}} + \mathbf{h}_{t-1}\mathbf{W_{hf}} + \mathbf{b_f}) \label{lstm2} \\
\mathbf{c}_t& = \mathbf{f}_t \cdot \mathbf{c}_{t-1} + \mathbf{i}_t \cdot tanh(\mathbf{x}_j \mathbf{W_{xc}} + \mathbf{h}_{t-1}\mathbf{W_{hc}}) \label{lstm3}\\
\mathbf{o}_t& = \sigma(\mathbf{x}_j \mathbf{W_{xo}} + \mathbf{h}_{t-1} \mathbf{W_{ho}} + \mathbf{b_o}) \label{lstm4}\\
\mathbf{h}_t& = \mathbf{o}_t \odot tanh(\mathbf{c}_t).
\end{align}


where the $\mathbf{W}$'s and $\mathbf{b}$'s are the parameters of the LSTM, $\mathbf{h}_t$ is a real-valued hidden-state vector at timestep $t$, $\sigma(\cdot)$ is a sigmoid function, and $\odot$ represents an element-wise multiplication. The whole network structure is described in Figure~\ref{fig:lstm-structure}. In the last timestep $T$, $\mathbf{h}_T$ encodes all the previous information of the sequence and can be used for classification. This $\mathbf{h}_T$ is fed to a softmax layer whose target is the class label $\mathbf{y} \in [0,1]^C$, associated with the input sequence. Assuming that the LSTM predicts $\hat{\mathbf{y}}=\textbf{encoder}(\mathbf{X})=$softmax$(\mathbf{h}_T)$, we will use the cross-entropy as the classification loss function:
\begin{align}
\mathcal{L}_1(\mathbf{X}, \mathbf{y}) & = - \mathbf{y} \log(\hat{\mathbf{y}} ) = - \mathbf{y} \log(\textbf{encoder}(\mathbf{X})).
\end{align}

The $\textbf{encoder}(\cdot)$ encodes the whole sequence into a hidden representation. The encoder can be realized in may ways, such as RNN, or (Attention) LSTM.

Theoretically, a sufficiently large and well-trained neural network model should be able to perform sequence classification perfectly~\cite{neubig2017neural}, because neural networks are universal function approximators~\cite{gybenko1989approximation}. However, in practice, it is necessary to learn these functions from limited data.
A drawback of standard LSTM for classification is the long-distance dependencies problem. For example, the beginning of signals might have less impact on the decision. The LSTM memory cell are designed to mitigate this problem by allowing the LSTM memory cells to store and access information over long periods of time. Unfortunately, due to the noisy or irrelevant signals, 
it is hard to guarantee that we will learn to handle these properly.

\subsection{LSTM with Temporal Attention}
\begin{figure}
\centering
\includegraphics[width=0.35\textwidth]{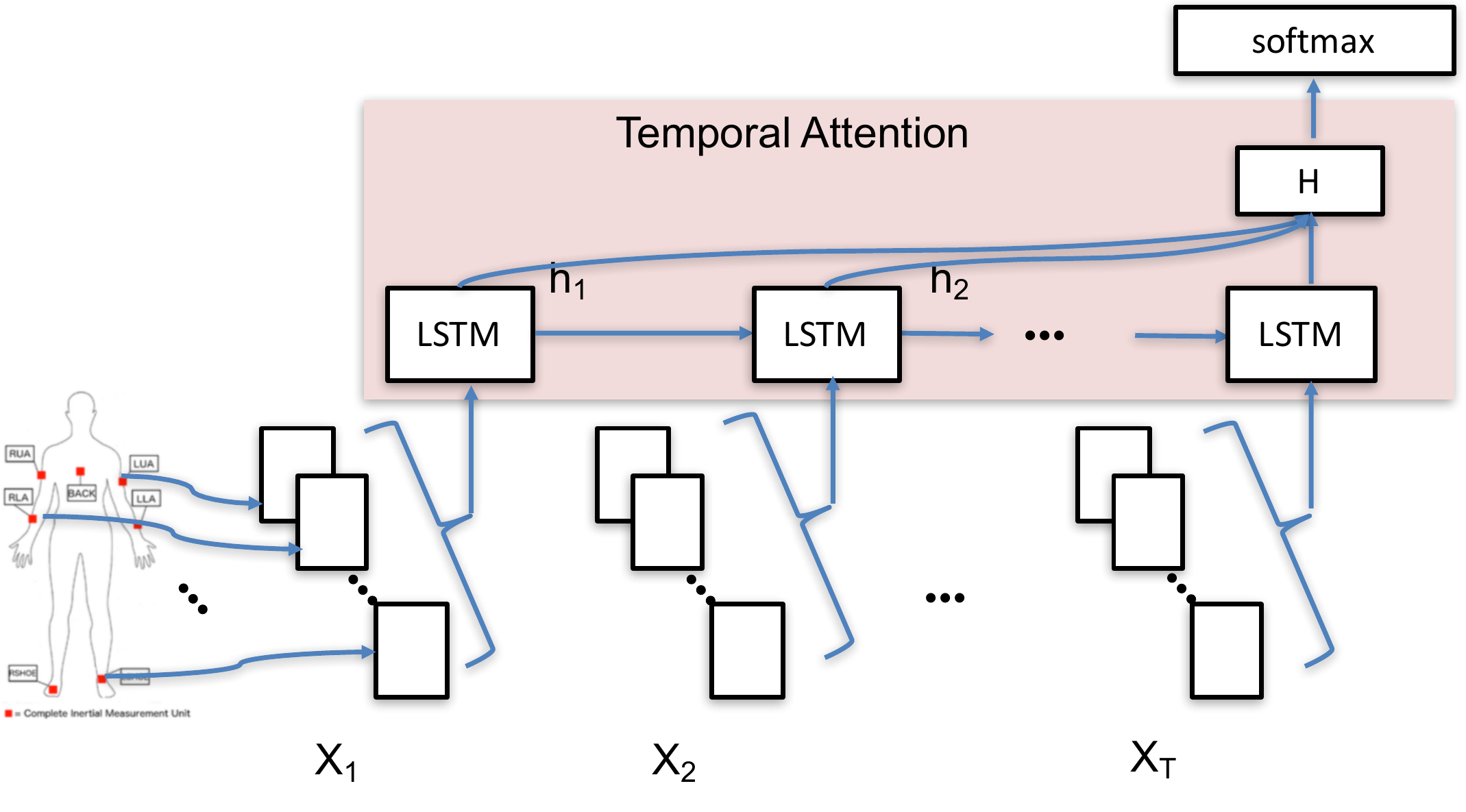}
\caption{\label{fig:time-att-structure} Structure of our LSTM + Temporal Attention for HAR}
\end{figure}

The basic idea of temporal attention is that instead of attempting to learn a single vector representation
for the whole signal in the last timestep, we instead keep around vectors for every timestep in the input signal. These vectors are referenced at the final step for classification (Figure~\ref{fig:time-att-structure}). As a result, we can express input signal in a more efficient way. To compute the vector $\mathbf{H}$, instead of using the last hidden state vector $\mathbf{h}_T$ in the LSTM~\cite{hochreiter1997long}, we consider a weighted sum of all the previous timesteps, 
\begin{align}
\mathbf{H} = \sum_{t=1}^T \alpha_t \mathbf{h}_t,
\end{align}
where the $\alpha_t$ are the attention weights. In standard LSTM~\cite{hochreiter1997long}, the $\alpha_T$ is fixed to be 1, and $\alpha_t=0$ when $t<T$. In the attention model, $\alpha_t$ are computed with a feed-forward neural network:
\begin{align}
\alpha_t = \frac{\exp\{\text{score}(\mathbf{h}_T, \mathbf{h}_t)\}}{\sum_{s=1}^T \exp\{\text{score}(\mathbf{h}_T, \mathbf{h}_s)\}}.
\end{align}
Here, the $\text{score}(\cdot)$ is a bilinear function:
\begin{align}
\text{score}(\mathbf{h}_t, \mathbf{h}_s) = \mathbf{h}_t^T \mathbf{W}_{\alpha} \mathbf{h}_s,
\end{align}

\begin{figure*}[tbph]
\centering
\includegraphics[width=0.5\textwidth]{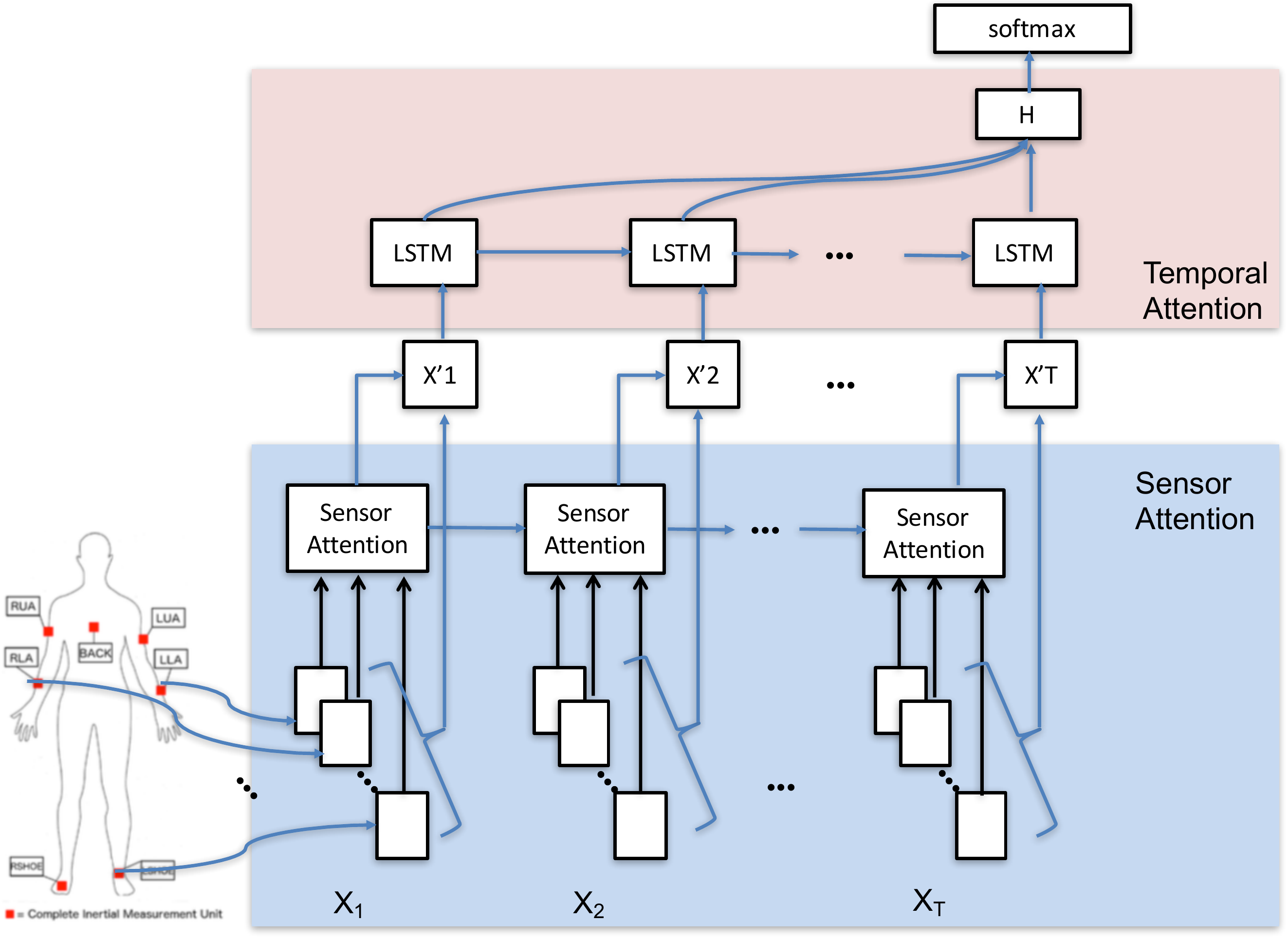}
\caption{\label{fig:time-sensor-att-structure} Structure of our novel Attention-based LSTM (Temporal Attention + Sensor Attention) framework for HAR. From the bottom, the signals coming from the multiple wearable sensors are re-weighted by attention LSTM, which also allows to capture time series information. On top of that, one more LSTM layer is stacked to capture more abstract features. The top layer is a softmax classifier.}
\end{figure*}

where, $\mathbf{W}_{\alpha}$ is a matrix of learnable parameter. In this way, the model is able to revisit the previous information and focus on more important parts to learn a better representation. The corresponding loss function is given by
\begin{align}
\mathcal{L}_2(\mathbf{X}, \mathbf{y}, \boldsymbol{\alpha}) = - \mathbf{y} \log(\textbf{encoder}(\mathbf{X}, \boldsymbol{\alpha})).
\end{align}



\subsection{LSTM with Sensor Attention}
We would also like to capture the potentially varying importance of sensor modalities. To do so, we propose a sensor attention mechanism on the input layer. Different from the temporal attention, the sensor attention weights are calculated based on two different information sources: (i) previous attention history $\boldsymbol{\beta}_{t-1}$ and (ii) current input signal $\mathbf{x}_t$ (Figure~\ref{fig:time-sensor-att-structure}). An unnormalized scalar “energy” value, $\mathbf{e}_t$, is produced for each memory entry:
\begin{align}
& \mathbf{e}_t = \mathbf{w}_e^T \tanh(\mathbf{W}_{\beta} \boldsymbol{\beta}_{t-1} + \mathbf{W}_x \mathbf{x}_t)\\
& \boldsymbol{\beta}_t = \frac{\exp(\mathbf{e}_t)}{\sum_{c=1}^C \exp(\mathbf{e}_t^c)}\\
& \mathbf{x}^{\prime}_t = \boldsymbol{\beta}_t \odot \mathbf{x}_t.
\end{align}

Here, $\mathbf{x}_t^{\prime}$ is the reweighted signal input at time $t$ and $\odot$ is element-wise multiplication. The loss function is given by
\begin{align}
\mathcal{L}_3(\mathbf{X}, \mathbf{y}, \boldsymbol{\alpha}, \boldsymbol{\beta}) = - \mathbf{y} \log(\textbf{encoder}(\mathbf{X}, \boldsymbol{\alpha}, \boldsymbol{\beta})). \label{eq:time-sensor-att}
\end{align}

Note that $\alpha_t$ at timestep $t$ is a scalar for reweighting the hidden representations, while $\boldsymbol{\beta}_t$ is a vector whose dimension equals the number of channels\footnotemark[1]. In standard LSTM~\cite{hochreiter1997long}, the sensor weight can be regarded as uniform distribution.

\footnotetext[1]{We have different granularity to apply the attention mechanism, e.g. on sensors, on sensor modalities, or on each channel for each modality. But the smaller granularity requires more data to train the model. Here we only focus on sensor modality, and use sensor attention and sensor modality attention interchangeably.}


\subsection{Improving Attention LSTM with Continuous Attention} 
One problem of the previous attention approaches is that the distribution of the attention weights is sharp. This is reasonable for NLP, because the discrete tokens in a sentence independently convey meanings. For example, in the sentence ``This is a good restaurant,'' the word ``good'' shows a positive sentiment. However, when we recognize the activity from a series of signals, the selected signals should be a consecutive series rather than represent discrete, disconnected points. We therefore introduce additional regularization terms for temporal attention and sensor attention respectively.

\subsubsection{Continuous Temporal Attention}
The continuous temporal attention regularization encourages continuity, which is given by
\begin{align}
\Omega_T(\boldsymbol{\alpha}) = \lambda_1 \sum_t|\alpha_t - \alpha_{t-1}|, \label{eq:continue-time-att}
\end{align}
where the regularization forces the continuous attention.  


\subsubsection{Continuous Sensor Attention}
The continuous sensor attention regularization discourages transitions. It is given by
\begin{align}
\Omega_S(\boldsymbol{\beta}) = \lambda_2 \sum_t|\boldsymbol{\beta}_t - \boldsymbol{\beta}_{t-1}|, \label{eq:continue-sensor-att}
\end{align}

where the regularization discourages switching sensor modalities back and forth. 

The final loss function is the combination of equation (\ref{eq:time-sensor-att}), (\ref{eq:continue-time-att}) and (\ref{eq:continue-sensor-att}), $\mathcal{L}(\mathbf{X}, \mathbf{y}, \boldsymbol{\alpha}, \boldsymbol{\beta}) = \mathcal{L}_3(\mathbf{X}, \mathbf{y}, \boldsymbol{\alpha}, \boldsymbol{\beta}) + \Omega_T(\boldsymbol{\alpha}) + \Omega_S(\boldsymbol{\beta})$. Because the attention weights are not provided during training, we actually minimize the expected loss
\begin{align}
\label{eq:final_loss}
\min_{\theta} \sum_{(\mathbf{X,y}) \in D} \mathcal{L}(\mathbf{X}, \mathbf{y}, \boldsymbol{\alpha}, \boldsymbol{\beta}),
\end{align}

where $\theta$ denotes the set of learnable parameters in the model, and $D$ is the collection of training instances. Our continuous attention objective (\ref{eq:final_loss}) encourages the model to compress the input signal into coherent representations that work well for the recognition. To minimize the loss, we can use mini-batch gradient descent.

\section{Experiments}
We selected three publically available HAR datasets for our evaluation. All datasets reflect human activities in different contexts and have been recorded by various sensors ({\em e.g.}, accelerometers, gyroscope, etc.). Sensors were either worn or embedded into objects that subjects manipulated. The sensor data was segmented using a sliding window as described. All the machine learning experiments were carried out on a server equipped with a Tesla K20c GPU and 64G memory. 

Training details include: 1 layer LSTM with 128 dimension hidden representation; optimization approach is ADMA with 0.05 learning rate; gradient normalization at 1.

\subsection{Datasets and Setup}
We use three publicly available datasets in our experiments with the same settings as previous works. The first is the \textbf{PAMAP2} dataset (Physical Activity Monitoring for Aging People 2)~\cite{reiss2012introducing}. It consists of 12 lifestyle
activities (``walking,'' ``lying down,'' ``standing'' etc.) and sport exercises (``nordic walking,'' ``running,'' etc.) by 9 participants. Accelerometer, gyroscope, magnetometer,
temperature, and heart rate data are recorded from inertial measurement units (IMUs) located on the hand, chest and ankle over 10 hours, resulting in 52 dimensions. To compare the result with previous work~\cite{guan2017ensembles, hammerla2016deep}, we downsampled the
data from 100Hz to 33.3Hz, and used a 5.12 second sliding window with 78\% overlap, resulting in around 473k samples. All samples were standardized to zero mean and unit variance. As in previous work~\cite{hammerla2016deep, guan2017ensembles}, we use Participant 6 for test, Participant 5 for validation, and the rest of the participants for training.

The second dataset used for our experiment is the \textbf{Daphnet Gait (DG)} dataset~\cite{bachlin2009potentials}. It contains recordings of 10 Parkinson' disease (PD) patients instructed to perform activities that are likely to induce freezing of gait. Freezing of gait (FOG) is common in advanced PD, where affected patients struggle to initiate movements such as walking. The goal is to detect these freezing incidents. This is a binary classification problem. Accelerometer readings were recorded from ankle, knee, and trunk, resulting in 9 dimensions. We use Participant 9 for validation, Participant 2 for test, and the rest of the data for training. We use the same settings as used before~\cite{hammerla2016deep}, and downsampled the data to 32Hz. The sliding window size is 1 second, resulting in around 470k samples for training.


The third dataset used is the \textbf{Skoda Mini Checkpoint (Skoda)} dataset~\cite{zappi2008activity}, which describes the activities of assembly-line workers in a car production scenario. The dataset contains a worker wearing 19 accelerometers on both arms while performing 46 activities in the factory at one of the quality control checkpoints. To recognize the right arm's gestures (``checking the boot,'' ``opening engine bonnet,'' etc.) in our experiments, we focus on 10 accelerometers placed on the right arm. The recording is about 3 hours long, consisting of 70 repetitions per gesture with 98Hz sampling rate. We downsampled the data to 33Hz, and standardized to zero mean and unit variance, resulting in 60 dimensions and around 190k samples. We use 80\% of the of the data in each class for training, the next 10\% for validation, and the rest for test.

We use $F$-measure ($F_1$) in the evaluation. Since the traditional $F_1$ score is used to measure the performance of binary classification, we used mean F1 score ($F_m$) by weighting classes according their sample proportion
\begin{equation}
F_m = \frac{1}{C} \sum_{i=1}^{C} \frac{2 \cdot \text{Precision}_i \cdot \text{Recall}_i }{\text{Precision}_i + \text{Recall}_i},
\end{equation}
where for the given class $i$,
$\text{Precision}_i = \frac{\text{TP}_i}{\text{TP}_i+\text{FP}_i}, \text{Recall}_i = \frac{\text{TP}_i}{\text{TP}_i+\text{FN}_i}.$
Here, $i=1,...,C$ is the set of classes considered, $\text{TP}_i$,$\text{FP}_i$ represents the number of true and false positive, respectively and $\text{FN}_i$ is the number of false negatives.

\begin{table*}[h!]
\centering
\begin{tabular}{|l|l|l|l|}
\hline
\textbf{Models}                & \textbf{PAMAP2} & \textbf{DG} & \textbf{Skoda}\footnotemark[4] \\ \hline
LSTM baseline (\cite{hochreiter1997long}) (LSTM without Attention)            &  $0.7548$& $0.6675$ &$0.9040$       \\ \hline
DeepConvLSTM (\cite{ordonez2016deep})     &  $0.7480$&  $0.7344$\footnotemark[3] &$0.9120$       \\ \hline
LSTM-S (\cite{hammerla2016deep})       &  $0.8820$& $0.7600$   &$0.9210$      \\ \hhline{|=|=|=|=|}
LSTM + Temporal Attention                    &$0.8052$& $0.7913$  &$0.9240$          \\ \hline
LSTM + Sensor Attention                &$0.7384$& $0.6700$  &$0.9002$           \\ \hline
LSTM + Continuous Temporal Attention        &$0.8629$& $0.8216$  &$\boldsymbol{0.9381}$            \\ \hline
LSTM + Continuous Sensor Attention      &$0.7797$& $0.7817$  &$0.8802$           \\ \hline
LSTM + Continuous Temporal + Continuous Sensor Attention\footnotemark[5]  &$\boldsymbol{0.8996}$& $\boldsymbol{0.8373}$   &$0.8903$           \\ \hline
\end{tabular}
\caption{Comparison of recognition results achieved using our attention-based LSTMs (bottom five result rows) versus baseline using the state-of-the-art (top three result rows) for sample-wise activity recognition (mean F1 score). Our LSTM + Continuous Temporal Attention is significantly better than the LSTM baseline with $p$-value $< 0.01$.}
\label{table:main_result}
\end{table*}

\begin{figure}[t!]
   \centering
  \includegraphics[width=0.4\textwidth]{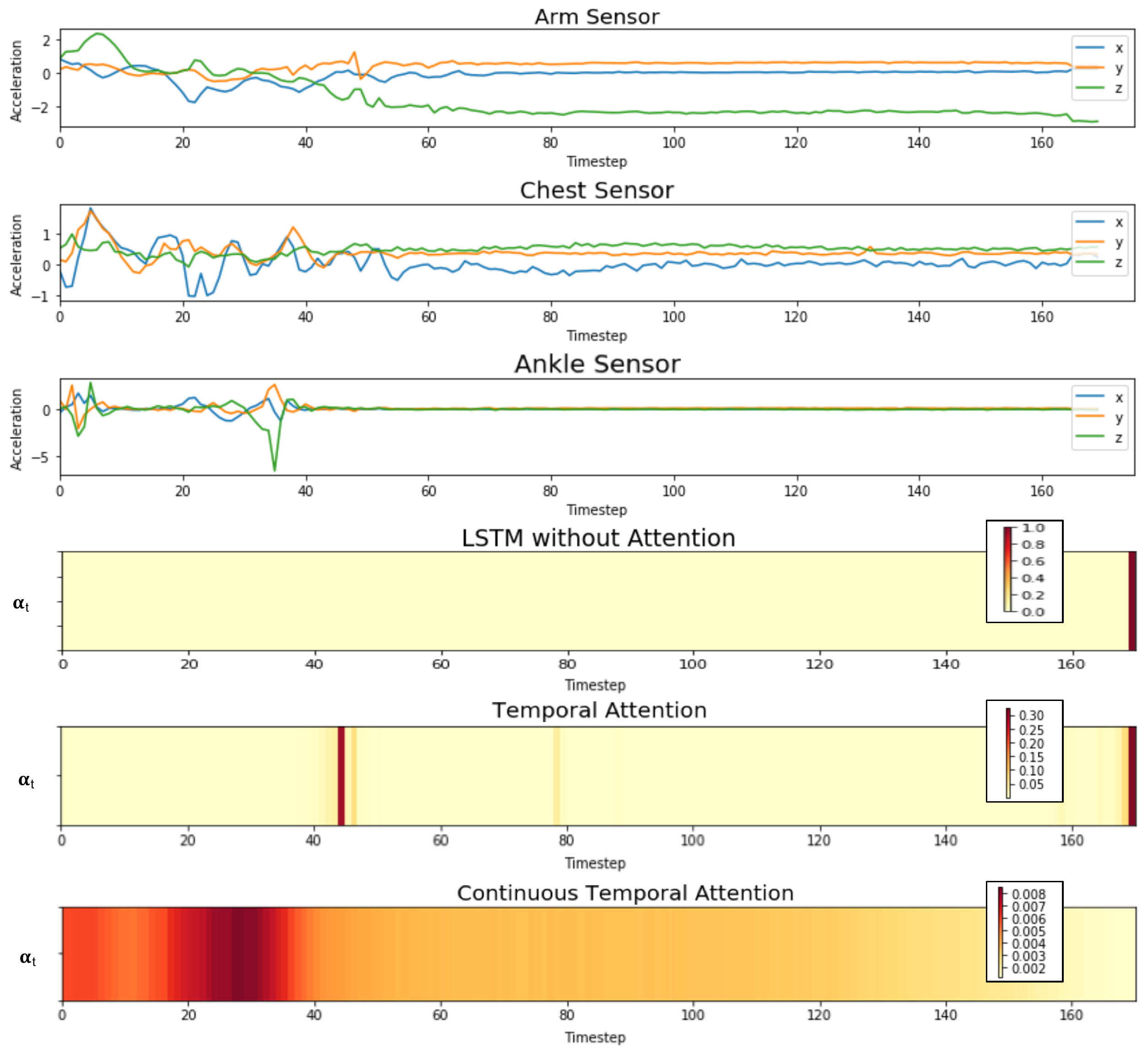} 
  \caption{An example of a 5.12s window labeled as ``walking'' activity on PAMAP2. However, only the beginning of the acceleration signal shows it is walking. The 4th, 5th and 6th rows show the visualization of attention weights for LSTM without attention, LSTM + Temporal Attention and LSTM + Continuous Temporal Attention, respectively.}\label{fig:att_1}
\end{figure}

\subsection{Comparing with Traditional RNN Models for HAR}
We compare our attention-based models to state-of-the-art DNN models for HAR. The baseline methods include standard LSTM~\cite{hochreiter1997long}, DeepConvLSTM~\cite{ordonez2016deep}, and LSTM-S (LSTM + sample-by-sample analysis)~\cite{hammerla2016deep}. Although our models also perform better than the ensemble LSTM~\cite{guan2017ensembles}\footnotemark,[2] it is fair to compare with single model. In this paper, we focus on the evaluation on the single models.

\footnotetext[2]{On PAMAP2 and Skoda, our models achieve higher mean F1 scores, compared to the results of the ensemble LSTM reported in ~\cite{guan2017ensembles}.}

The results are in Table~\ref{table:main_result}. Our LSTM Continuous Temporal Attention outperforms the state-of-the-art, LSTM-S, on DG and Skoda. Specifically, the Continuous Temporal Attention achieves around $14.50\%$, $17.00\%$, and $3.40\%$ improvements in mean F1 scores on the three datasets, compared to the LSTM baseline~\cite{hochreiter1997long}.

Combining Continuous Sensor Attention with Continuous Temporal Attention can further improve the mean F1 scores on PAMAP2 and DG datasets with around $1.8\%$ and $7.7\%$, compared to the state-of-the-art. It is interesting that the continuous sensor attention has a negative impact for Skoda. This might be because recognizing the right arm activities in car assembly lines depends on the the sensor interaction rather than a single sensor Thus constantly focusing on a single sensor, which is done in continuous sensor attention might deteriorate the classification performance on Skoda.

\begin{figure}[t!]
   \centering
  \includegraphics[width=0.4\textwidth]{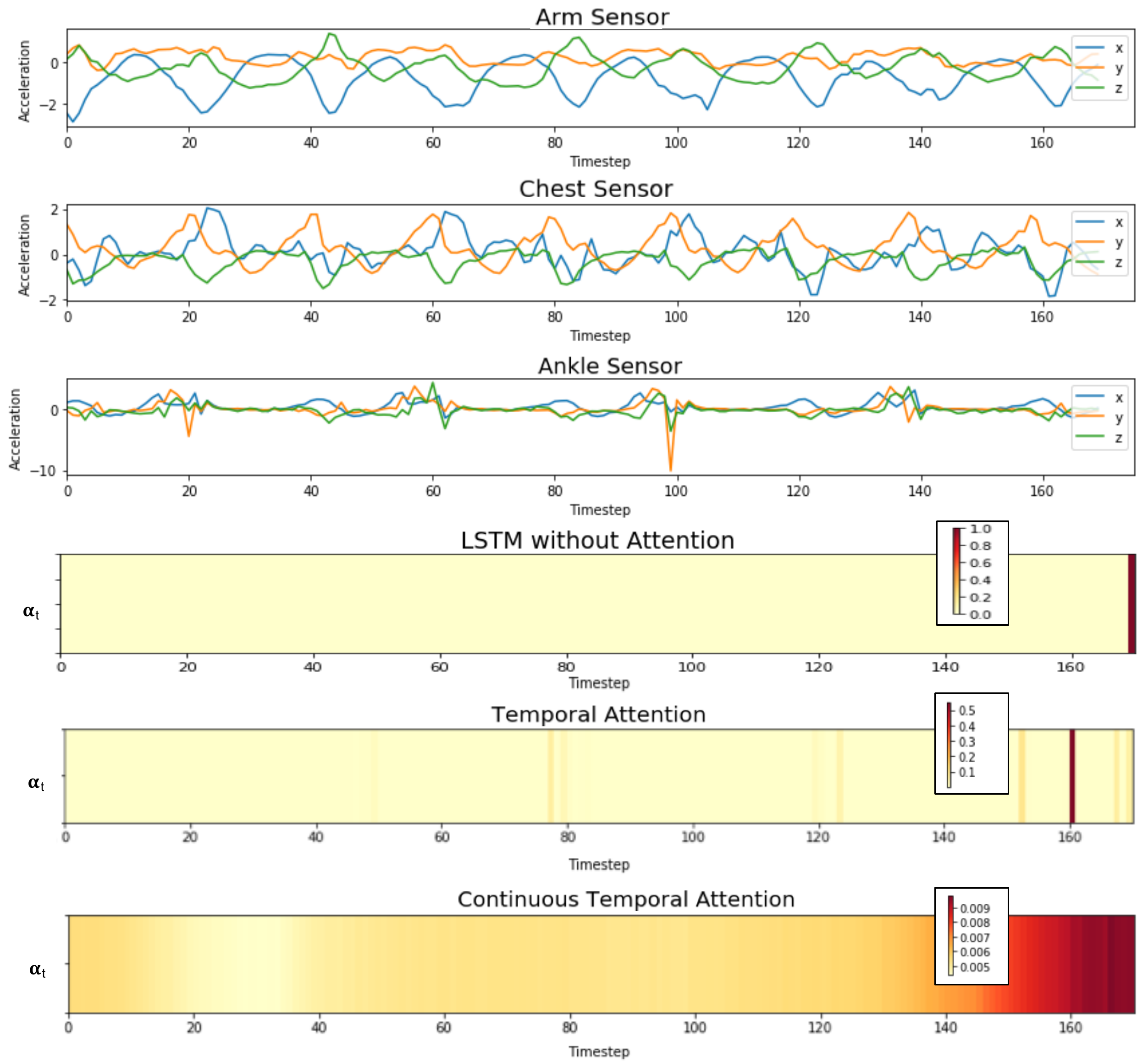} 
  \caption{An example of a 5.12s window labeled as ``walking'' activity on PAMAP2. The entire signal shows the pattern of walking. The 4th, 5th and 6th rows show the visualization of attention weights for LSTM without Attention, LSTM + Temporal Attention and LSTM + Continuous Temporal Attention respectively.}\label{fig:att_2}
\end{figure}

\subsection{Visualizing Important Signal Components}


This section seeks to better understand how temporal attention models help achieve high mean F1 scores in HAR. We now study which are the important signal components by looking into the temporal attention weights of the models.

\begin{figure}[t!]
   \centering
  \includegraphics[width=0.4\textwidth]{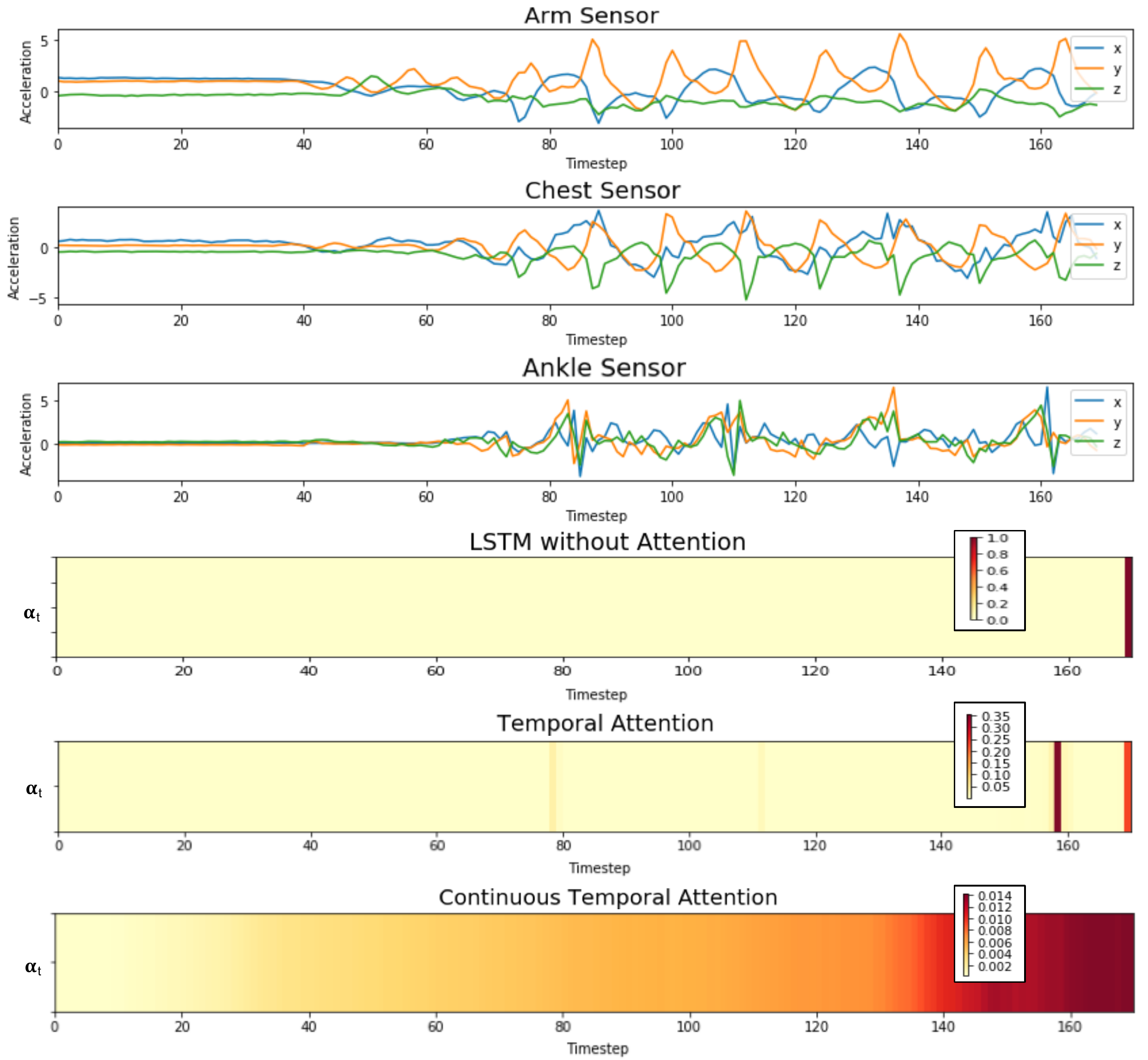} 
  \caption{An example of a 5.12s window labeled as ``running'' on PAMAP2. Only the last half of the acceleration signal reflects it is running. The 4th, 5th and 6th rows show the visualization of attention weights for LSTM without Attention, LSTM + Temporal Attention and LSTM + Continuous Temporal Attention, respectively.}\label{fig:att_3}
\end{figure}

We visualize the attention weights from the LSTM baseline (LSTM without attention), the temporal attention model and the continuous temporal attention model for two different walking activities(Figure~\ref{fig:att_1},~\ref{fig:att_2}) and one running(Figure~\ref{fig:att_3}) on PAMAP2 dataset. Because the accelerometers are generally important to recognize daily-life activities~\cite{bao2004activity}, we only show the raw acceleration signals (the first three rows) from the arm sensor, the chest sensor, and the ankle sensor. The forth and fifth rows are the attention weights for each timestep. The darker color indicates the higher weight.

Figure~\ref{fig:att_1} shows a signal window labeled as ``walking'', but only the beginning has the ``walking'' signal. The standard LSTM will use the last encoded hidden vector, $\mathbf{h}_T$, for recognition, which contains many irrelevant stationary signals. The temporal attention is able to look back at timestep 43, which is the end of the walking signal, and it encodes the previous information. Because the last hidden vector ($\mathbf{h}_T$) still contains some ``walking'' information, the attention model also uses the last timestep for recognition. In contrast, the continuous temporal attention model is able to force the model to attend the hidden vectors for walking signals (around timestep 0-43) and its attention weight decays to the time after walking (around timestep 70).

\footnotetext[3]{We implemented DeepConvLSTM~\cite{ordonez2016deep}, got the similar result for PAMAP2, and applied it on DG dataset. The rest mean F1 scores are from previous works.}
\footnotetext[4]{The result of Skoda is from~\cite{guan2017ensembles}.}
\footnotetext[5]{We choose the optimal value of $\lambda_1=0.1$ and $\lambda_2=0.5$.}

Figure~\ref{fig:att_2} shows the attention weights for a window labeled as ``walking.'' We observe some clear patterns in the entire raw signals, such as period and amplitude. The temporal attention model highlights the hidden vector close to the end of the signal, while the continuous temporal attention attends the hidden vector towards the end consecutively.
Due to the recurrent structure, the hidden vector close to the end is more informative than its beginning
Thus if the entire signal segment contains the target activity, the temporal attention model performs similar to the standard LSTM. Nevertheless, it is interesting to notice that the continuous temporal attention puts less attention between timestep 20 - 40, which corresponds to the flat signal from the ankle sensor. 

Figure~\ref{fig:att_3} shows the attention weights a window labeled as ``running'', but different from Figure~\ref{fig:att_2}, only the last half of the window shows ``running'' behavior. Even though the LSTM representation encodes the irrelevant components (static signals) at the beginning, the hidden vectors close to the end accumulate more information than the previous ones. The temporal attention tends to attend the hidden vectors towards the end, but it also pay some attentions in the middle. Starting from timestep 80, the continuous temporal attention gradually increases its interest in the hidden vectors and focuses more in the end.


 

\subsection{Visualizing Important Sensor Modalities}
\begin{figure}[h]
   \centering
  \includegraphics[width=0.5\textwidth]{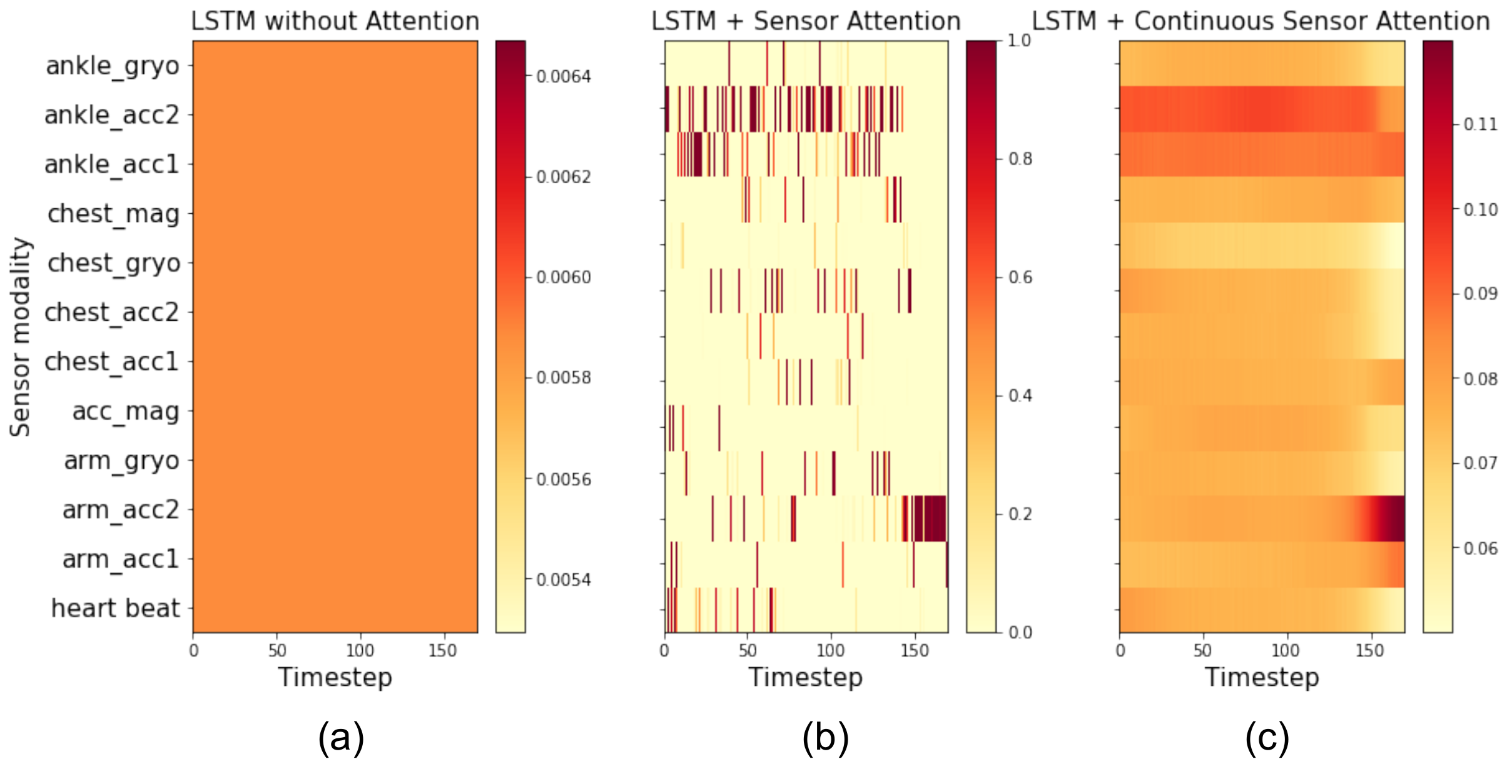} 
  \caption{The visualization of LSTM (a) without sensor attention, (b) sensor attention and (c) continuous sensor attention, for a walking activity on PAMAP2 dataset. LSTM without attention is identical to the uniform attention weight.}\label{fig:att_sensor}
\end{figure}

We now study the impact of different sensor modalities placed on different parts of users body. We compare the attention weights from sensor modality attention and continuous sensor modality attention in Figure~\ref{fig:att_sensor} (b) and (c). Both of the sensor modality attention approaches put high emphasis on the ankle sensor (Acc2, Gyro) and the arm sensor (Acc2). This is more reasonable than the LSTM without attention mechanism, which treats all the sensor modality equally. In the figure, The sensor attention is shifting-focus, acting as a feature selector in each timestep (Figure~\ref{fig:att_sensor} (a)). The continuous sensor attention has smooth attention weights. This can prevent the model switching between sensors for recognition (Figure~\ref{fig:att_sensor} (b)).

\section{Discussion}
In this section, we discuss some issues of our attention-based approaches and the potential improvements for future work.

\textbf{Dealing with sensor interaction}
Our sensor attention model uses the current corresponding raw signal and previous attention state to infer the current attention state. To better involve the previous attention state and previous signal, we can in future work use another LSTM to model the sensor attention state transition. In this way, the LSTM sensor attention is able to capture more complicated sensor interactions. 

\textbf{Better understanding for temporal attention}
We rely on the attention-based LSTM to interpret the recognition process. However, the signal information accumulates through time in LSTM, which makes it hard to determine how much data from each timestep contribute to the classification.
Although LSTM is able to forget the unimportant signals and remember the important signals, to quantitatively evaluate the impact from the previous signals and determine signal importance, In future work, we need to use other approach, such as analyzing the gradients during the back propagation~\cite{leino2018influence}. In addition, there is mismatching between attention and continuous attention in some cases (e.g. Figure~\ref{fig:att_1}), more analysis can be done to understand the models' behaviors.

\section{Conclusion}

We propose temporal attention, sensor attention, and two continuous constraints to understand and improve recurrent networks for human activity recognition. The attention-based models are able to focus on salient signal components and important sensor modalities. The experimental results demonstrate that our proposed approaches can achieve improvements in mean F1 score compared to the state-of-the-art. With visualization results, we show that our approaches improve the interpretability of the deep neural networks for HAR. 


\section{Acknowledgments}\label{sec:acknow}
We thank Anupam Datta, John P. Shen, Zhike Mao, and the anonymous reviewers for helpful comments and valuable discussion. This research is supported in part by the National Science Foundation under the award 1704845 and research fundings from Ericsson and Intel.


\bibliographystyle{acm-sigchi}
\bibliography{bibliography}

\end{document}